\documentclass[conference]{IEEEtran}
\IEEEoverridecommandlockouts

\usepackage{cite}
\usepackage{amsmath,amssymb,amsfonts}
\usepackage{algorithmic}
\usepackage{graphicx}
\usepackage{textcomp}
\usepackage{amsmath}
\usepackage{array}
\usepackage{multicol}
\usepackage{rotating}
\usepackage{adjustbox}
\usepackage{multirow}
\usepackage{makecell}
\usepackage{xcolor}
\usepackage{tabularx} 
\usepackage{longtable}
\usepackage{array}
\usepackage{float}
\usepackage{caption}
\usepackage{booktabs} 
\usepackage{multirow} 
\usepackage{siunitx} 
\def\BibTeX{{\rm B\kern-.05em{\sc i\kern-.025em b}\kern-.08em
    T\kern-.1667em\lower.7ex\hbox{E}\kern-.125emX}}
\begin{document}

\title{Intelligent Systems in Neuroimaging: Pioneering AI Techniques for Brain Tumor Detection \\
}

\author{\IEEEauthorblockN{1\textsuperscript{st} Md. Mohaiminul Islam}
\IEEEauthorblockA{\textit{Dept.of ECE} \\
\textit{North South University}\\
Dhaka, Bangladesh \\
mohaiminul.islam7@northsouth.edu}
\and
\IEEEauthorblockN{2\textsuperscript{nd} Md. Mofazzal Hossen}
\IEEEauthorblockA{\textit{Dept.of ECE} \\
\textit{North South University}\\
Dhaka, Bangladesh \\
mofazzal.hossen@northsouth.edu}
\and
\IEEEauthorblockN{3\textsuperscript{rd} Maher Ali Rusho\textsuperscript{*}}
\IEEEauthorblockA{
NMR Spectroscopist\\
Lassonde School of Engineering York University \\
Toronto, Canada \\
alirusho@yorku.ca}
 \and
 \IEEEauthorblockN{4\textsuperscript{th} Nahiyan Nazah Ridita}
 \IEEEauthorblockA{\textit{Dept.of ECE} \\
 \textit{North South University}\\
 Dhaka, Bangladesh \\
 nahiyan.ridita@northsouth.edu}
 \and
 \IEEEauthorblockN{5\textsuperscript{th} Zarin Tasnia Shanta}
 \IEEEauthorblockA{\textit{Dept.of ECE} \\
 \textit{North South University}\\
 Dhaka, Bangladesh \\
 zarin.shanta@northsouth.edu}
 \and
 \IEEEauthorblockN{6\textsuperscript{th} Md. Simanto Haider}
 \IEEEauthorblockA{\textit{Dept.of ECE} \\
 \textit{North South University}\\
 Dhaka, Bangladesh \\
 simanto.haider@northsouth.edu}
 \and
 \IEEEauthorblockN{7\textsuperscript{th} Ahmed Faizul Haque Dhrubo}
 \IEEEauthorblockA{\textit{Dept.of ECE} \\
 \textit{North South University}\\
 Dhaka, Bangladesh \\
 ahmed.dhrubo@northsouth.edu}
 \and
 \IEEEauthorblockN{8\textsuperscript{th} Md. Khurshid Jahan}
 \IEEEauthorblockA{\textit{Dept. of ECE} \\
 \textit{North South University}\\
 Dhaka, Bangladesh \\
 khurshid.jahan@northsouth.edu}
 \and
 \IEEEauthorblockN{9\textsuperscript{th} Mohammad Abdul Qayum}
 \IEEEauthorblockA{\textit{Dept.of ECE} \\
 \textit{North South University}\\
 Dhaka, Bangladesh \\
 mohammad.qayum@northsouth.edu}
 }

\maketitle

\begin{abstract}
This study deliberates on the application of advanced AI techniques for brain tumor classification through MRI, wherein the training includes the present best deep learning models to enhance diagnosis accuracy and the potential of usability in clinical practice. By combining custom convolutional models with pre-trained neural network architectures, our approach exposes the utmost performance in the classification of four classes: glioma, meningioma, pituitary tumors, and no-tumor cases. Assessing the models on a large dataset of over 7,000 MRI images focused on detection accuracy, computational efficiency, and generalization to unseen data. The results indicate that the Xception architecture surpasses all other were tested, obtaining a testing accuracy of 98.71\% with the least validation loss. While presenting this case with findings that demonstrate AI as a probable scorer in brain tumor diagnosis, we demonstrate further motivation by reducing computational complexity toward real-world clinical deployment. These aspirations offer an abundant future for progress in automated neuroimaging diagnostics. 
\end{abstract}


\section{Introduction}
Nowadays, Brain tumor is a severe problem in this world. Every year, many people are affected by this disease, and people die from brain tumors. The cost of this disease treatment is very high. People who belong to the upper middle class and high class can bear the cost of this treatment, but people who belong to the lower middle class or lower class cannot bear the cost of this disease treatment. Every year with over 300,000 cases reported annually on a worldwide basis, brain tumors are a consistently pressing concern for the international medical community\cite{b1}.

A brain tumor is a growth of cells in or near the brain. It can happen in the brain tissue or near the brain tissue. There is a possibility that cancer spreads to the brain from other parts of the body, which is called the secondary brain tumor\cite{b2}. Brain tumors proliferate more than other tumors of the human body. Sometimes, these brain tumors are found when they are tiny because they cause symptoms you notice immediately. Other brain tumors grow very large before they're discovered\cite{b2}.

There are many different types of brain tumors, but benign brain tumors tend to be slow-growing brain tumors. On the other side, malignant brain tumors tend to be fast-growing brain tumors. It is estimated that the incidence of brain tumors in general is around 2-3 cases per 100,000 individuals in Bangladesh\cite{b3}. In developing countries like India, complete registration of brain tumors and reliable data collection rarely occur due to monetary constraints. Male patients are more affected than female cases except in meningioma\cite{b4}. Brain tumors have bimodal age distribution with a peak in childhood and adult age groups of 45–70 years\cite{b5}. Clinical presentation of brain tumors depends on the location, size of the tumors, and growth rate of the neoplasm.

As per the WHO classification, CNS tumors have extensive classification and subtypes. Glial tumors are the most common type of brain tumor and include astrocytoma, ependymoma, glioblastoma, oligodendroglioma, and others\cite{b6}. In Bangladesh, in general, all these specialists are not available in one location, which presents considerable difficulty for clinicians and patients in treating brain tumors. Because of various reasons in Bangladesh, a large number of patients did not begin treatment or were lost to follow-up after initiation of therapy\cite{b6}. For the person who continued treatment, considerable inconvenience was caused because of multiple parallel doctor appointments and therapy sessions in various departments of a single hospital. Lack of collaboration was most evident in challenging cases and uncommon tumors that lacked uniformity in decision-making. An increased need existed for interaction and coordination between the treating specialists, significantly improving the overall quality of service.

\section{Literarture Review}

Early and accurate diagnosis of brain tumors is key for the treatment of the patient and, therefore, numerous studies have investigated the use of artificial intelligence (AI) and deep learning (DL) in medical imaging to tackle this challenge.

In this respect, Abdusalomov et al. have showcased success in transfer learning by fine-tuning the YOLOv7 model for the detection of brain tumors. A remarkable estimation of 99.5\% accuracy was achieved using a comprehensive dataset of 10288 images, which included gliomas, meningiomas, pituitary tumors, and non-tumor cases; this study revealed the transition of learning with state-of-the-art results\cite{b7}. In a novel pipeline for MRI-based brain tumor detection, Anantharajan et al. combined an Adaptive Contrast Enhancement Algorithm (ACEA), fuzzy C-means segmentation, and hybrid ensemble deep neural support vector machine classifier. Their approach achieved excellently balanced accuracy (97.93\%), sensitivity (92\%), and specificity (98\%)\cite{b8}. Mathivanan et al. took advantage of transfer learning with architectures like ResNet152, VGG19, DenseNet169, and MobileNetv3 for the diagnostic purpose of brain tumors. Of all architectures, they attained the highest accuracy of 99.75\% on the MobileNetv3 model, indicating the effectiveness of lightweight architectures in high-performance medical imaging\cite{b9}. Sudipto Paul et al. reportedly employed YOLOv5 for brain tumor segmentation. They emphasized the efficiency of the model concerning runtime and computational cost without sacrificing accuracy\cite{b10}. Saeedi et al. similarly adopted a 2D Convolutional Neural Network (CNN) and auto-encoder in tumor detection from MRI, whereby an augmented dataset was utilized to enhance classification accuracy\cite{b11}.

Among these findings, some claim that classical machine learning methods, especially ones combined with deep learning, have turned out to be quite versatile. Some examples would include biologically inspired wavelet transforms and feature-extraction methods such as the Gray Level Co-occurrence Matrix (GLCM) combined with genetic algorithms, for segmentation and classification of the tumor region\cite{b12}. Tseng and Tang optimized the eXtreme Gradient Boosting (XGBoost) method in an effort to identify brain tumors reliably, by means of image preprocessing and selecting important features\cite{b13}. They changed the contrast of the images by using Contrast-Limited Adaptive Histogram Equalization (CLAHE) and clustered the images using the K-Means algorithm. Their XGBoost model was able to classify images with an accuracy of 97\%. They also tested Naïve Bayes and ID3 classifiers. Amin et al. \cite{b14}proposed a method in which the deep features were extracted using the InceptionV3 model. They took the score vector out of the softmax layer and used it as input to the Quantum Variational Classifier (QVR) to differentiate between glioma, no tumor, meningioma, and pituitary tumor. The model was assessed using three datasets, Kaggle,2020-BRATS, and a locally captured image set, which were all established datasets and acute detection accuracies were all achieved above 90\%. The InceptionV3 architecture consists of 315 total layers with 94 convolutional layers, 94 ReLU layers, 94 batch normalization layers, 4 max-pooling layers, 9 average pooling layers, 1 global pooling layer, 1 fully connected layer, 1 output classification layer, 15 depth concatenation layers, and a softmax for final classification. Anil and Rajesh introduced a machine learning-based technique for recognizing and classifying tumorous versus non-tumorous regions in brain MRI images\cite{b15}. Their methodology included feature extraction, segmentation, classification, and data preprocessing. The active tumor was segmented using the Chan-Vese (C-V) segmentation technique by selecting an appropriate starting point. Performance was evaluated using the Dice Coefficient (DC), the Jaccard Similarity Index (JSI), and the Structural Similarity Index (SSIM). Their SVM-based classifier performed better than KNN, achieving an accuracy of over 98\%.

Paul and his team used transfer learning techniques to retrain a model using MRI slices split into three groups\cite{b16}. Gradient-weighted Class Activation Mapping (Grad-CAM) was used to visualize the network's attention during prediction. The achieved categorical accuracy was 0.93. Rajesh, Malar, and Geetha\cite{b17} presented a process of extracting features using an accurate computer-aided diagnostic system of MRI brain images for tumor classification. Rough Set Theory was applied for feature extraction, a PSONN was trained, and therefore tested, using the MRI brain images in order to classsify the tumors into normal and abnormal categories. A Data Adaptive Filter was introduced to effectively remove impulse noise from the images. The optimization process was carried out using Particle Swarm Optimization, with the PSONN acting as a supervised learning model. Kharrat et al.\cite{b18} applied the Wavelet Transform for image segmentation by decomposing MRI images. Specifically, they used the Daubechies Wavelet, which yielded optimal results in their tests. The unsupervised K-Means algorithm was also employed. Their methodology enhanced contrast using mathematical morphology, thereby reducing the steps required for feature extraction. Rafia et al. \cite{b19}proposed several approaches for brain tumor detection on MRI images. They evaluated their model on a dataset called the Brain Tumor Figshare (BTF). They evaluated a number of methods, however, it was discovered that the best detection model was the combination of YOLOv5 to detect and a 2D U-Net to produce a pixel-wise tumor segmentation mask. For the test data, the YOLOv5 algorithm had mean Average Precision (mAP) of 89.5\% (the maximum) while the combined YOLOv5 + 2D U-Net combination had a higher Dice Similarity Coefficient (DSC) compared to only using 2D U-Net. 

Abhishek et al.\cite{b20} using the OASIS and BRATS2018 datasets included image descriptions of images in terms of images being tumor and non-tumor. Of the entirety of images of greater than 20,000 total, they used 80\% for supervised training and 20\% for testing. They implemented transfer learning simulation using two models: AlexNet and VGGNet using MATLAB 2018b's Deep Learning Toolbox. Of the two models, the VGG19 model achieved the highest accuracy. Tanjim et al.\cite{b21} used a Convolutional Neural Network (CNN) and a transfer learning model to classify the three types of brain tumors. They used an ensemble method that grouped together many pre-trained models to achieve high accuracy. An implementation of a CNN and VGG-16 model had a validation accuracy of 0.97687, and a test accuracy of 0.9801. Naik and Patel conducted a Decision Tree classification algorithm and tested it against the Naive Bayes classfiier\cite{b22}. For brain tumor classification, the proposed presented Decision Tree model surpassed the Naive Bayes classifier. The Decision Tree produced an accuracy of 96\% and a sensitivity of 93\%. Amin \& his team presented an unsupervised clustering approach for tumor segmentation\cite{b23}. The model was report using multiple MRI modalities including DWI, FLAIR, T2, T1, \& T1c - their process focused on lesion enhancement, lesion segmentation, feature extraction, and classification. Performance evaluation was conducted using 5-fold cross-validation without overfitting bounded by 0.5 holdout. 

Saeed et al.\cite{b24} presented a new method to early identification of brain tumors and cerebrospinal fluid (CSF) accumulation. Their hybrid method based on K-Nearest Neighbour (K-NN) algorithm, Fast Fourier Transform (FFT) and Laplace Transform demonstrated early diagnosis using four-dimensional (4D) MRI data. In their method they utilized Light Editing Field (LEF) tools leveraging a range of attributes; spatial values, objects properties, cancer stages, and demographic traits they applied their K-means. Khaliki and Basarslan\cite{b25} studied CNN and several models of transfer learning, prototype depth, InceptionV3, EfficientNetB4, and VGG19. In evaluating individual performance they asked for metrics such as F-score, recall, precision, and accuracy. Their maximum performance increased under the VGG16 model, which had an accuracy of 98\%, F-score of 97\%, AUC of 99\%, Recall of 98\%, and precision of 98\%.

Irsheidat and Rehab\cite{b26} developed their model based on Convolutional Neural Networks (CNN) that examined magnetic resonance images. They did this from a mathematical form and specified standard operations from matrices. The model achieved a validation accuracy of 96.7\% and a test accuracy of 88.25\%. The segmentation and shape extraction process included Binary Thresholding, Edge Detection, High-Pass Filtering, ToZero Thresholding, and filtering using other templates. Devkota et al. \cite{b27} proposed a multi-modulal fuzzy potential segmentation technique (MMFPS). While each of the components in their pipeline had benefits as compared to standard techniques, the difficulty was at the segmentation phase using sFCM algorithm. Their model had a cancer detection accuracy of 92\% and 86.6\% accuracy on the classifier, although they did not as a factor of accuracy class between cancer stages (Stage I–IV).

Kumar et al. \cite{b28}utilized a multi-step pipeline with pre-processing, segmentation and classification with Support Vector Machine (SVM) and Artificial Neural Networks (ANN), which was explicitly expanded to include 3D reconstruction of regions with a tumor. SVM was responsible for the classification while the ANN learned discrete tumor features. Also included in methodology near the classification stage are PDE based filtering and segmentation using a signed distance map. This method is more involved than the basic SVM-only pipeline. All of these research papers show the huge potential of AI for medical imaging. Challenges still exist in terms of computational complexity, large and heterogeneous datasets, and trustworthiness in the real world. Our study expands upon these works by incorporating the Xception architecture and a custom CNN model with the aim of achieving a balance between accuracy, generalization, and computational efficiency. 
 
\section{Background Study}
\subsection{Current Research}

Recent advances in artificial intelligence and deep learning have revolutionized the accuracy and efficiency of brain tumor detection and classification within MRI images. Of these approaches, Convolutional Neural Networks have consistently gained traction due to a general reliance on superior extracting capabilities of the patterns. The Convolutional Neural Networks make use of a hierarchical arrangement of layers in acquiring vital features such as edges, textures, and other tumor-specific traits. The efficiency is enhanced yet further thanks to pooling layers that reduce the spatial dimension, while classification is performed by fully connected layers\cite{b29}. U-Net is a CNN architecture devised specifically for image segmentation, which has been found quite effective for the segmentation of tumor regions harnessing an encoder-decoder structure, thus making pixel-wise segmentation possible\cite{b30}. The recurrent neural networks and long short-term memory have also been applied to sequential data analysis, particularly for time-series data or multi-slice MRI frames\cite{b31}.

Other models like Deep Belief Networks (DBN) and Support Vector Machines have demonstrated predictability in tumor detections, exploiting feature hierarchy and binary classification power, respectively\cite{b32,b33}. Traditional ensemble methodologies, correction applied including Random Forest, have been applied in the feature-classification task, showcasing versatility in combining hand-crafted features with automated detection\cite{b34}. Novel methods like Generative Adversarial Networks (GANs) enabled synthetic data generation for dataset augmentation for better generalization of models across varying imaging cases\cite{b35}. Transfer learning using models such as ResNet, DenseNet, and Inception-based architectures have further catalyzed progress allowing models trained on other domains to adapt quickly to medical image tasks\cite{b36}.

Notwithstanding the advances, features such as computational burden, overfitting, and lack of generalisation on unseen data remain topical. Much work utilizes the current research to make attempts to address these issues by improving model architecture, using diverse datasets, and improving training pipelines. We expand on this work by integrating Xception architecture and a custom CNN model; this aims to improve classification accuracy while ensuring computational efficiency. 

\subsection{Our Approach}

To overcome the challenges of precise and accurate brain tumor detection, we develop a hybridized approach that takes advantage of pre-trained architectures and a custom-designed CNN model. We, however, stress accuracy above all else with low computational complexity so that the concept can be realized in real-life clinical applications.

Taking exceptional feature extraction into consideration, the Xception model on Imagenet.com was chosen as one of today's topmost architectures of deep learning. It employs lightweight depthwise separable convolutions without compromising accuracy, thus reducing computational overhead. The model was pre-trained with weights from a TensorFlow tensorflow.keras.applications\cite{b37} library, and training was done depending on the dataset to tailor the model to the specific classification task. More layers, including fully connected layers and dropout layers at 50 and 25 percent, respectively, were thereafter added to enhance generalizability and minimize overfitting. The final layer is, thus, activated by the softmax function for a multi-class classification comprised of four classes: glioma, meningioma, no tumor, and pituitary.

Embedding proprietary CNN architecture for efficient classification became central for our experimentation. Our architecture consists of four convolutional blocks with growing filter sizes starting from 32, 64, 128, and finally 256. Each convolutional block has batch normalization and uses Leaky ReLU\cite{b38} activation to stabilize the network and avoid gradient vanishing. Max-pooling layers are engaged to minimize spatial dimensions and computational overheads. The further introduced architecture remains compact and adept at working with computationally limited environments; it further serves the great need as a real-time application.

Moreover, we could benchmark models like ResNet50, InceptionV3, and InceptionResNetV2 while comparing the best-performing models, namely the Xception model and the custom CNN model with these architectures and thereby establish their relative advantages based on performance in terms of accuracy, validation loss, and computational efficiency.

We thus propose to find this balance between performance and sense-making: making an effective solution for brain tumor detection realistically integrable into real-world clinical workflows. 

\section{Methodology}
\subsection{Data Analysis}
In this study, we, therefore, merged data from the Figshare, SARTAJ, and Br35H datasets to obtain a highly diverse dataset available at Kaggle\cite{b39}. The dataset consisted of 7,023 MRI images differentiated into four classes: glioma, meningioma, no tumor, and pituitary. The images portraying the non-tumor class were taken from the Br35H dataset specifically to provide good representations of non-pathological cases.

Additionally, all images were reshaped to be of equal dimension (224, 224, 3) for compatibility with deep learning architectures like Xception and a custom CNN model. Standardization of input dimensions ensured that features could be extracted uniformly across all models. Image normalization helped by bringing pixel values into a range between [0, 1], which in turn expedites the process of model training and helps with convergence.

Several data augmentation techniques (rotation, width, and height shift, shear transformations, zooming, and flipping) have been applied using the class ImageDataGenerator to increase generalization and assist with the reduction of overfitting, including: 
\begin{itemize}
 \item Rotating the images, as much as up to 40 degrees, to capture various angles in imaging.
 \item Shifting width/height by as much as 20\%, to account for positional variations.
 \item Employing shear transformations to account for geometric distortions.
 \item Zooming in and out from 20\% further, to imitate various levels of focal concentration.
 \item Flipping horizontally for duplicate mirror-image variations.
\end{itemize} 

The dataset was split into training, validation, and testing subsets, namely:
\begin{itemize}

 \item Taining Set: 70\% of the data for the training of models.
 \item Validation Set: 15\% of the data for hyper-parameter tuning and monitoring overfitting during training.
 \item Testing Set: 15\% of the data for assessing the final performance of models.

\end{itemize} 

\begin{figure}[htbp]
  \centering
  \begin{minipage}[b]{0.45\linewidth}
    \centering
    \includegraphics[width=\linewidth]{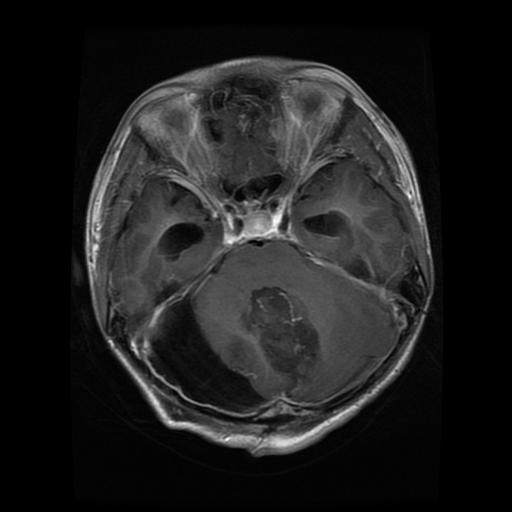}
  \end{minipage}
  \hfill
  \begin{minipage}[b]{0.45\linewidth}
    \centering
    \includegraphics[width=\linewidth]{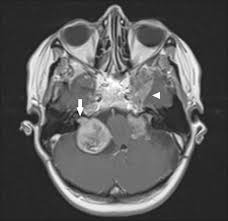}
  \end{minipage}
  
  \vspace{0.5cm}
  
  \begin{minipage}[b]{0.45\linewidth}
    \centering
    \includegraphics[width=\linewidth]{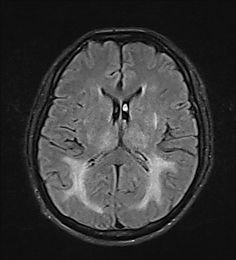}
  \end{minipage}
  \hfill
  \begin{minipage}[b]{0.45\linewidth}
    \centering
    \includegraphics[width=\linewidth]{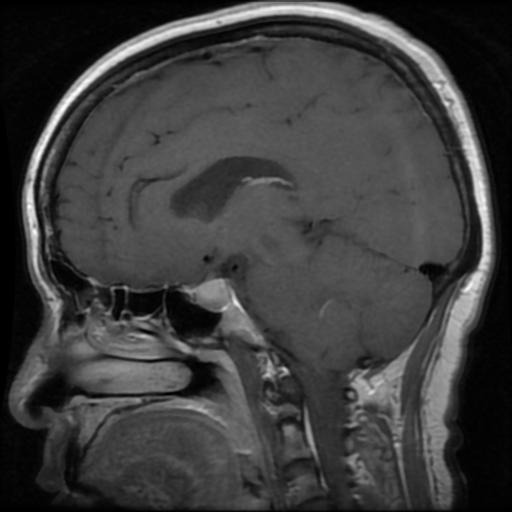}
  \end{minipage}
  \caption{MRI Scans Showing Various Conditions: Glioma, Meningioma,a Non-Tumor Case and Pituitary Tumor}
  \label{fig:comparison}
\end{figure}

A few images from each class were visually inspected to understand dataset balance and quality, with examples shown in Figure 1. The varying nature of gliomas, meningiomas, pituitary tumors, and non-tumor cases hence brought a firm rationale before the model building and evaluating stages.

By providing an amalgamation of extensive preprocessing, augmentative techniques, and a structured manner of splitting the datasets, this phase of the data analysis made sure the dataset was refined enough to train the models to very high accuracies with good generalization in brain tumor detection.

\subsection{System Design}

The system proposed in this paper relies on both the Xception model which aligns with a custom architecture for CNN to ensure effective and efficient brain tumor classification. On one hand, the Xception model implements depthwise separable convolutions for the purpose of efficient feature extraction with reduced computational complexity. Pre-trained weights from toothtensorflow.keras.applications library were fine-tuned on the given dataset, after which additional fully connected layers with dropout rates of 0.5 and 0.25 respectively were used to enhance generalization and avoid overfitting. A final softmax layer provided multi-class classification through the four categories: glioma, meningioma, no tumor, and pituitary.

In stark contrast lies the custom architecture designed from scratch, meant to suit situations wherein computational resource access is limited. It consists of four convolutional blocks characterized by convolutional layers, offers batch normalization, and applies Leaky ReLU activation to stabilize training and provide solutions to gradient vanishing problems. The first block had 32, the second 64, followed by 128 and 256 for the third and fourth respectively, while each block was followed by a max-pooling layer to curtail spatial dimension and computation costs. Loss value for multi-class classification is cross-entropy, and Adamax optimizer\cite{b40} at a fixed learning rate of 0.001 is adopted whereby the convergence is guaranteed to be stable and efficient. Here, Figure 2 is the system architecture of the custom CNN model that we use in our dataset for brain tumor detection.

\begin{figure*}[t]
\centering
\includegraphics[width=1\textwidth]{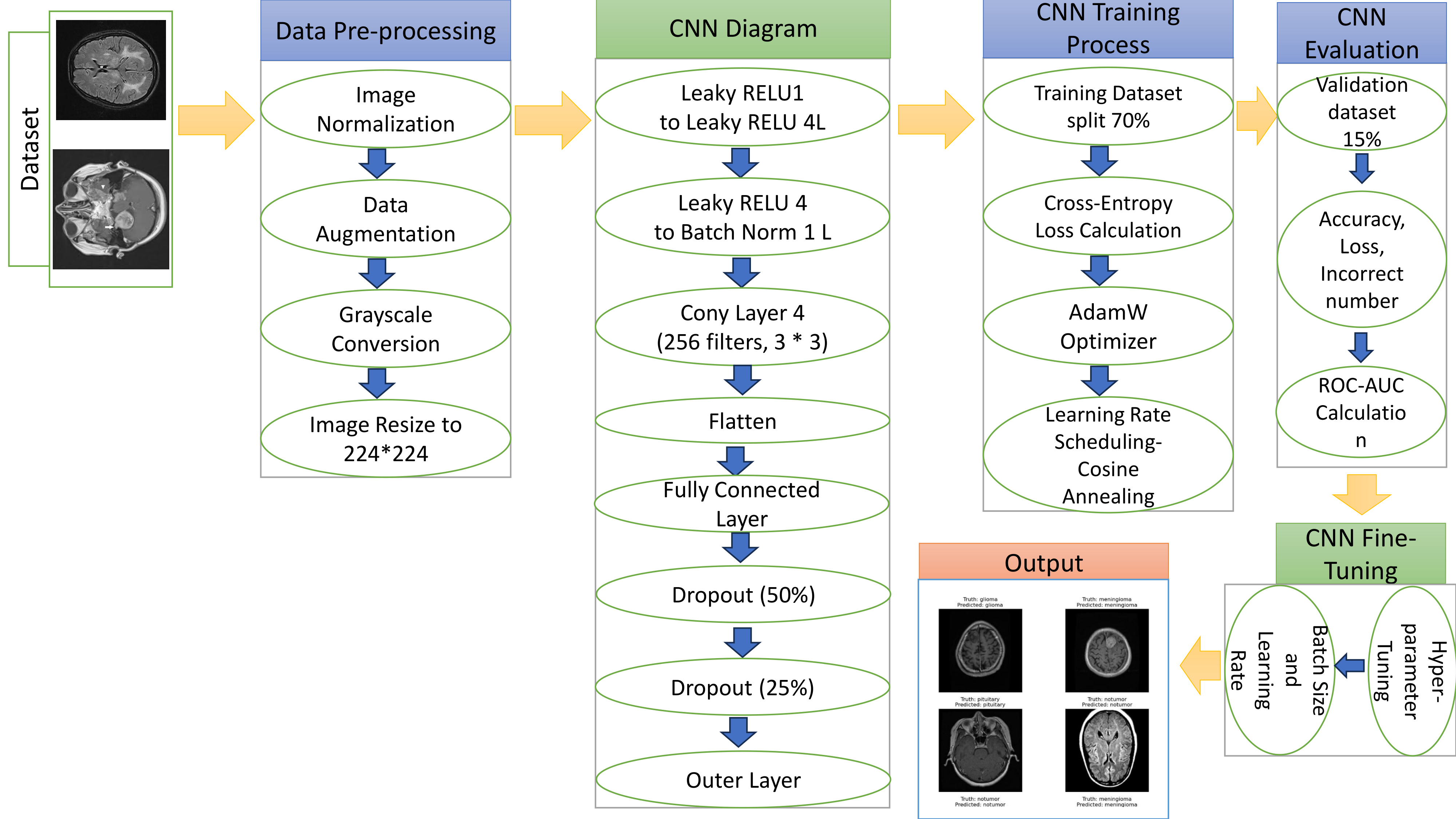}
\caption{Architecture of the proposed CNN system design.}

\end{figure*}

Lastly, in the training process, early stopping and model checkpoint mechanisms were also employed for further optimal purposes. Early stopping used validation loss, observing no improvements over their bygone five epochs, to stop training, preventing overfitting and thus unnecessary computation. Model checkpoints enabled the saving of the best model according to specific validation performance during training. The system was evaluated from all angles using metrics from confusion matrices, classification report, accuracy loss plots, allowing informed views into respective strengths and weaknesses of the models. The intersection of such design elements allows the system a well-compensated high degree of classification efficiency with very efficient computation, making it ideal in real-world clinical applications. 

\begin{figure*}[t]
\centering
\includegraphics[width=1\textwidth]{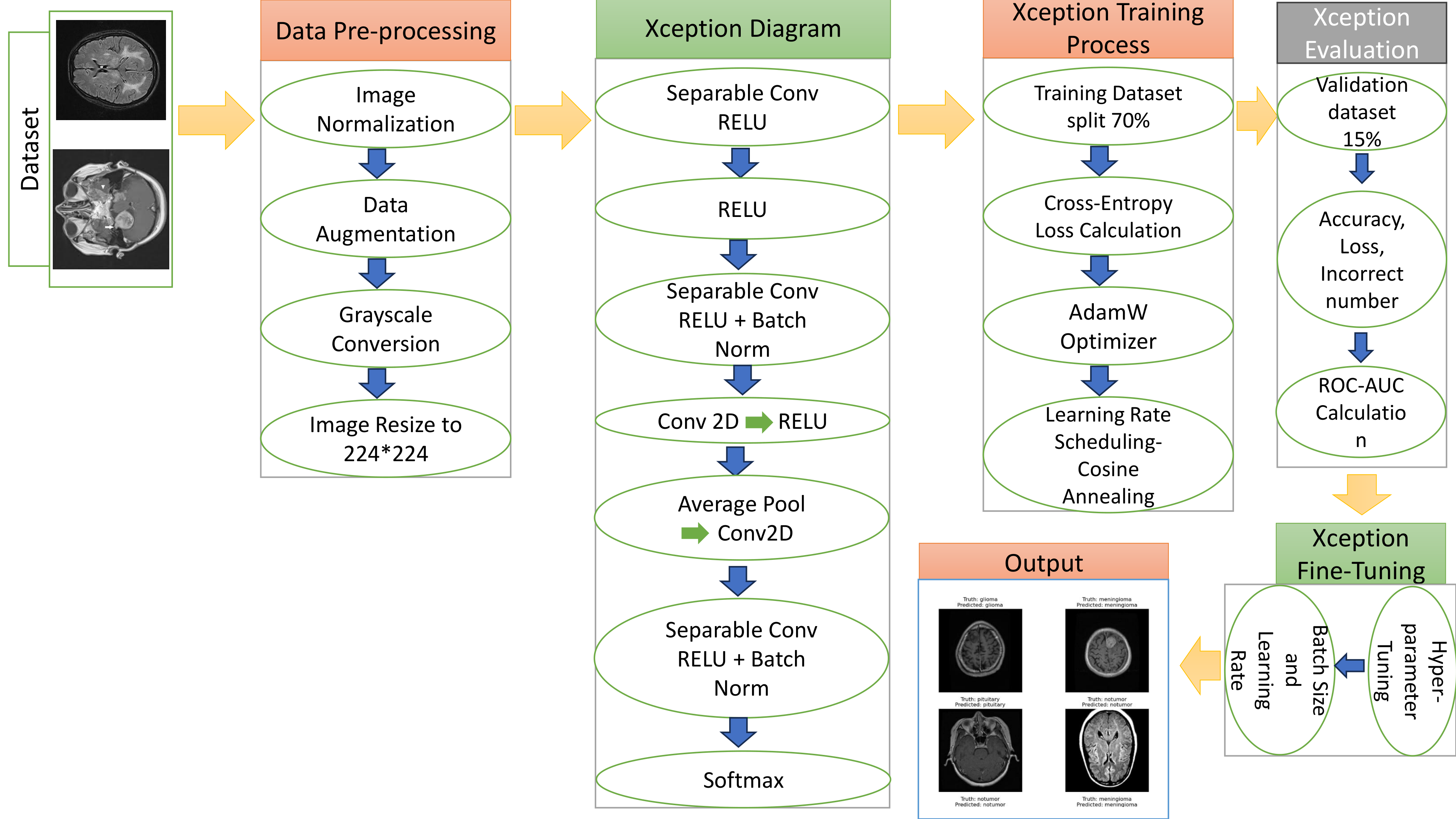}
\caption{Architecture of the proposed Xception system design.}
\label{fig:jv}
\end{figure*}

Here, Figure 3 is the system architecture of the proposed Xception model that we use in our dataset for brain tumor detection. We use Cross-Entropy Loss since this is a multi-class classification problem, and the definite cross-entropy loss function is used. The formula for categorical cross-entropy is as follows:

\[
L = - \sum_{i=1}^{N} \sum_{c=1}^{C} y_{i,c} \log(\hat{y}_{i,c})
\]

Here, N is the number of samples, C is the number of classes, which is the four classes of our dataset, y\_{i,c} is the actual label for the i-th sample and c-th class (one-hot encoded), ŷ\_{i,c} is the predicted probability for the i-th sample and c-th class, produced by the softmax function. The Adamax optimizer is a variant of the Adam optimizer and is used in this model. It is based on the infinity norm and is known for its stability with sparse gradients. Adamax works as follows:

\[
\theta_{t+1} = \theta_t - \alpha \frac{m_t}{v_t}
\]

Here, theta\_t are the parameters (weights) at time step t. Alpha is the learning rate (0.001 in this case). m\_t is the first-moment estimate (running mean of gradients). v\_t is the infinity norm of the gradient (max absolute value of gradient elements). Dropout is also used here to prevent overfitting in the model, and it works by randomly "dropping" neurons during training with a specified probability:

\[
h_i^{\text{dropout}} =
\begin{cases} 
h_i & \text{with probability } p \\
0 & \text{with probability } 1 - p 
\end{cases}
\]

Here, h\_i is the activation of the i-th neuron, and p is the dropout rate (0.5 and 0.25 in this model).

\section{Result and Analysis}

This research analyzed the classification of brain tumors based on MRI in four classes: glioma, meningioma, pituitary tumors, and no tumor. Several pretrained models along with a custom convolutional neural network were utilized for tumor classification. Performance metrics included training accuracy, testing accuracy, validation loss, and incorrect classification percentage. A summary of performance metrics can be found in Table I.
\begin{table}[ht]
\centering
\caption{Overall Model Performance Results}
\label{tab:1}
\begin{tabularx}{\columnwidth}{|*{11}{X|}} 
\hline
Model & Training Accuracy & Testing Accuracy & Training Loss & Valida- tion Loss & Incorr ect Predictions (Validation) & Percen- tage of Incorrect Predictions (Validation)  \\ \hline
Xcep tion & 99.11\% & 98.71\% & 0.0270 & 0.0496 48 & 18 & 1.258\% \\ \hline
ResNet 50 & 95.77\% & 94.80\% & 0.1383 & 0.9855 & 73 & 5.20\% \\ \hline
Incep tion V3 & 98.03\% & 98.00\% & 0.0736 & 0.1114 & 28 & 1.99\% \\ \hline
Incep tion ResNet V2 & 98.33\% & 98.14\% & 0.0495 & 0.0311 & 26 & 1.85\% \\ \hline
Custom CNN Model & 93.71\% & 91.64\% & 0.1862 & 0.2808 & 56 & 8.54\% \\ \hline
\end{tabularx}
\end{table}

The highest testing accuracy, recorded at 98.71\%, with low validation loss at 0.0496 was achieved by the Xception model, giving it an edge over the other models in terms of generalization and precision. With the least percentage of incorrect predictions (1.26\%) among all models tested, the robustness of the Xception model was reaffirmed. Figure 4 visually confirms the Xception model's capability in accurately classifying MRI images with minimal error.

InceptionV3 and InceptionResNetV2 models performed also strongly with testing accuracies of 98.00\% and 98.14\%, respectively, achieving comparable validation losses. Although these models were successful in feature extraction, Xception performed marginally better due to the lower rates of prediction error-it had incorrect predictions of 1.99\% and 1.85\%, respectively.

In contrast, the ResNet50 model fell short of other pre-trained models, showing testing accuracy at 94.80\% and validation loss at a discouraging 0.9855, reiterating the challenge of feature capturing from a complicated dataset. Although functional, the custom CNN with a testing accuracy of 91.64\% suffered from overfitting, a fact corroborated by a higher validation loss of 0.2808 and an incorrect prediction percentage of 8.54\%. Figure 4 represents the best model outcome, which predicts all MRI images with less error

\begin{figure}[htbp]
  \centering
  \includegraphics[width=1\columnwidth]{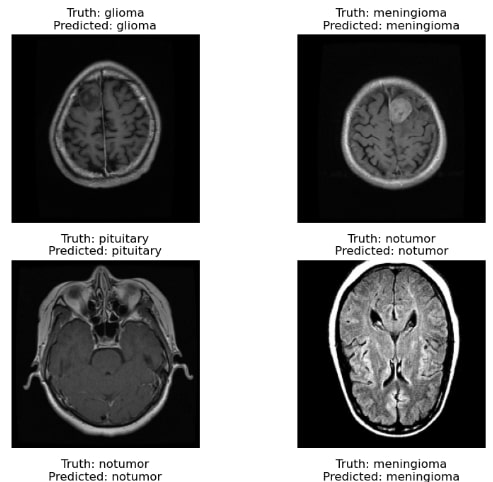} 
  \caption{Following the analysis of the MRI images, all detected brain tumors and no tumor were identified and classified.}

\end{figure}

\begin{figure}[htbp]
  \centering
  \includegraphics[width=1\columnwidth]{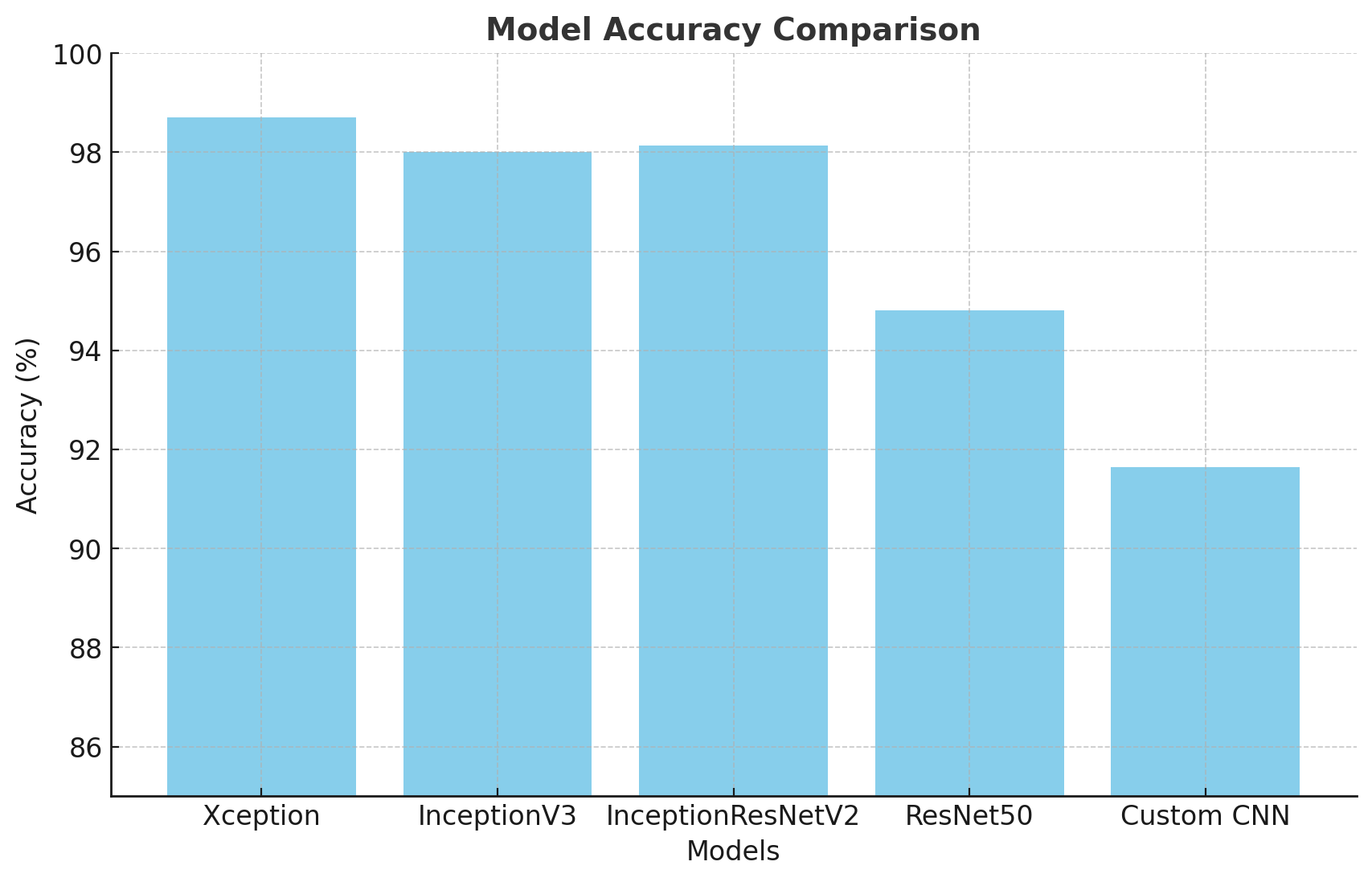} 
  \caption{Performance Evaluation Based on Graphical Representation of All Models.}
 
\end{figure}

Figures 5 illustrate the overall performance of each model, emphasizing the Xception model as the best-performing due to the balance of high accuracy, low validation loss, and generalization to unseen data. Further evidence for Xception being a strong classification model is reflected in the confusion matrix, showing an impressive number of class instances per class with totally accepted misclassifications.

\begin{figure}[htbp]
  \centering
  \includegraphics[width=1\columnwidth]{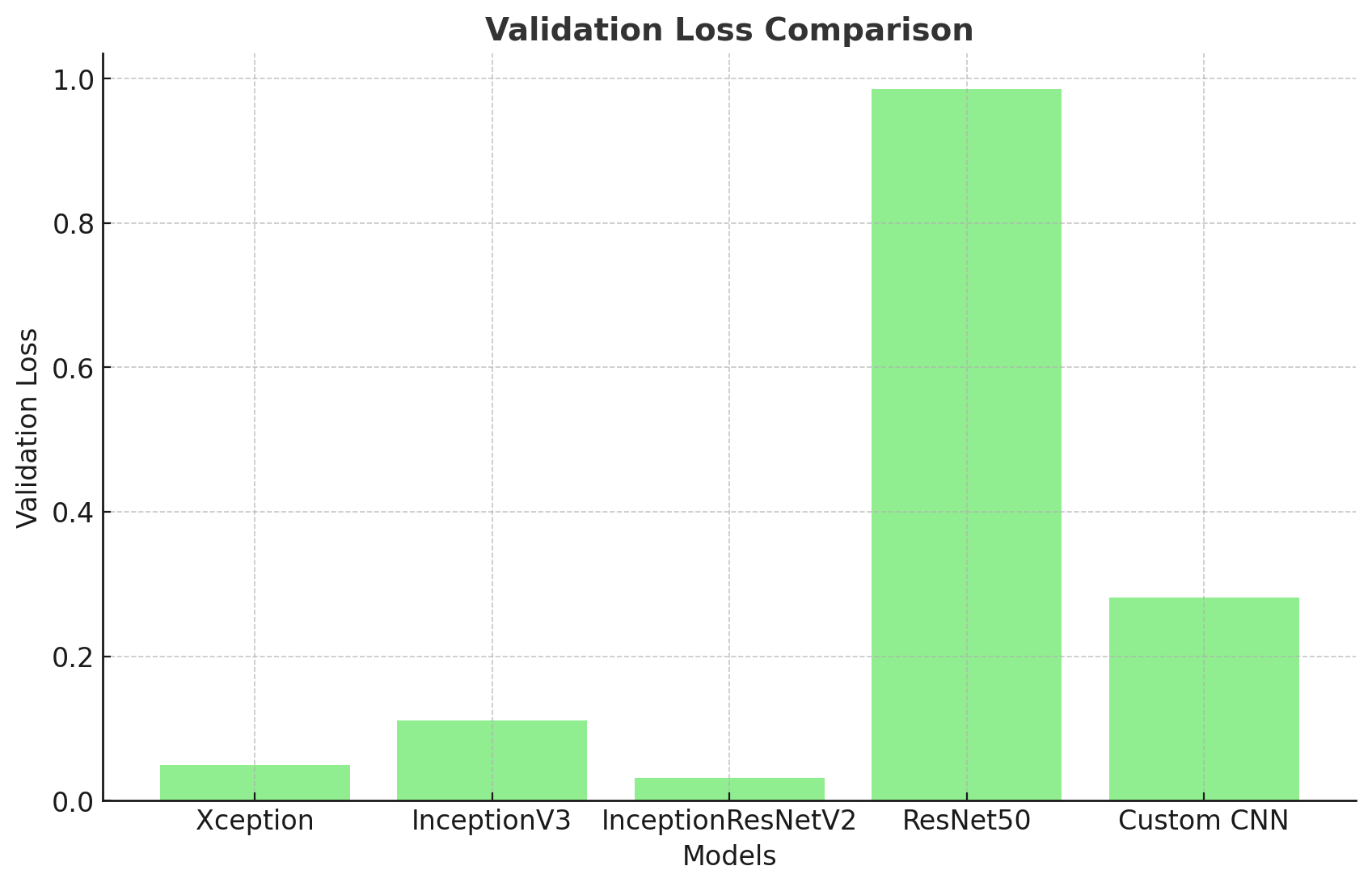} 
  \caption{Validation Loss Comparison Across Different Models.}

\end{figure}

\begin{figure}[htbp]
  \centering
  \includegraphics[width=1\columnwidth]{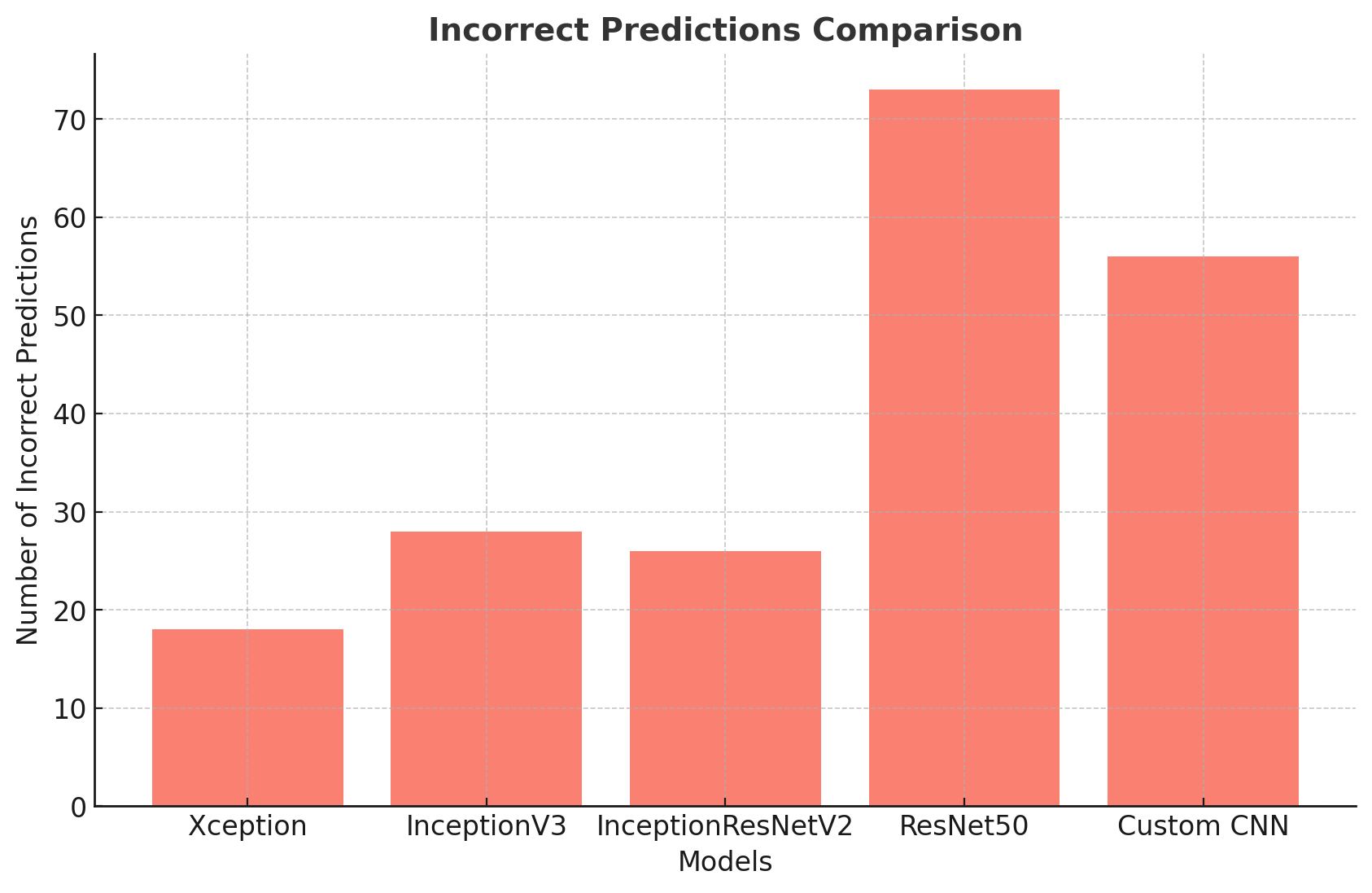} 
  \caption{Comparison of Incorrect Predictions Across Models.}

\end{figure}
Figures 6 \& 7 shows all models validation loss and incorrect predictions. These results demonstrate the effectiveness of the Xception model in using depthwise separable convolutions to extract complex features with computational efficiency. Its ability to compete with both traditional architectures and custom architectures highlights its potential for application in the real clinical world.

\section{Comparison with Other Papers}

Natha et al. \cite{b41}provide a timely and valid measurement of the need for accurate and efficient brain tumor identification, which is essential for appropriate treatment planning and ultimately the prognosis of patients. Natha et al. suggest an ensemble based deep learning approach that can be supplemented with various deep learning architectures to take advantage of each individual architectural strengths and avoid weaknesses. In this study they proposed different convolutional neural network (CNN) models such as  CNNs have performed well in image classification. The results from the experiments showed the SETL\_BMRI model to be an accurate and stable model with results revealing an overall accuracy of 98.70\%. In addition, an average precision, recall, and F1 score of 98.75\%, 98.6\% and 98.75\%, respectively. G´omez-Guzm´an et al. \cite{b42}study research proposes a CNN to automate brain tumor identification from MRI tense production. In this research, the CNN models used was a Generic CNN model, ResNet50, Inception V3, InceptionResNet V2, Xception, MobileNetV2, and EfficientNetB0. The research showed that when considering all CNN models including the generic CNN model as well as 6 pre-trained models, Inception V3 was the best CNN model for this dataset with an average Accuracy of 97.12

Güler \& Ersin \cite{b43}study explores several popular CNN architectures (VGG16, ResNet50, and InceptionV3) in terms of identifying brain tumor images. They take a methodological approach to tuning these models, tuning hyperparameters and using methods such as data augmentation to improve model generalizability. The main focus of their work is on determining the most effective CNN architecture in terms of accurately identifying and reliably detecting brain tumors. Here, VGG16, achieved 98.44\%; ResNet, achieved 98.81\%; InceptionV3, achieved the highest efficiency at 99.30\%, and Proposed Stack Ensemble Model at 99.66\%. Based on their findings the stack ensemble model, which captures the benefits from the three single CNN architectures, had the most accuracy which allows for improved brain tumor images classification. They also focused on different performance metrics, including: precision, F1-score, Matthews correlation coefficient (MCC), specificity, sensitivity, root mean square error (RMSE), mean squared error (MSE), mean absolute error (MAE), positive predictive value (PPV), negative predictive value (NPV), false positive rate (FPR), false negative rate (FNR); all suggesting their proposed model works well.

Saeed et al's \cite{b44}model is an ensemble of several deep learning architectures, including Convolutional Neural Networks and other advanced neural network models. An ensemble benefits from different architectures in order to produce a better classification capability than using each model individually. Each deep learning model is preprocessing the MRI images to extract features. The overall performance of the ensemble is better than using one model alone! The ensemble has better accuracy, precision, recall, F1 score that is again suggesting its potential in classifying brain tumors from MRI images. In using the ensemble method, Saeed et al's model had an accuracy of 92\%, precision of 90\%, recall of 92\%, and F1 score of 91\% on a 64 BS and 0.0001 LR.

\begin{table*}[t]
\centering
\caption{ANALYZING PERFORMANCE MEASURES IN COMPARISON TO OTHER TECHNIQUES}
\resizebox{\textwidth}{!}{%
\begin{tabular}{|p{3cm}|p{1.2cm}|p{4cm}|p{2cm}|p{5cm}|}
\hline
\textbf{Study} & \textbf{Year} & \textbf{Methodology} & \textbf{Accuracy} & \textbf{Notable Contributions} \\
\hline
Natha et al. & 2024 & Stacked ensemble model combining VGG19, InceptionV3, and DenseNet121 with majority voting & 98\% & Introduced an ensemble method to improve classification robustness \\
\hline
Gómez-Guzmán et al. & 2023 & Evaluated seven CNN architectures & 97.12\% & Thorough comparison of CNNs for brain tumor classification \\
\hline
Güler \& Ersin & 2024 & Optimized deep learning classifiers for brain tumor detection & 85\% & Automated system to assist radiologists and reduce manual effort \\
\hline
Saeed et al. & 2024 & Compared 10 DCNN models (EfficientNetB0, ResNet50, etc.) & 92\% & Ensemble model improved classification accuracy \\
\hline
Sandeep et al. & 2024 & Used pre-trained models adapted for brain tumor diagnosis & 99.75\% & Showcased the power of deep transfer learning in medical diagnosis \\
\hline
Zahoor \& Saddam & 2022 & Proposed Res-BRNet (residual + regional CNN) & 98.22\% & Outperformed traditional CNNs in robustness and accuracy \\
\hline
Kumar et al. & 2024 & Applied six ML algorithms & 99\% & Used LIME to improve interpretability in healthcare models \\
\hline
Krishnan et al. & 2024 & Used rotated patch embeddings for rotation invariance & 98.6\% & Used Grad-CAM for transparent decision explanation \\
\hline
\end{tabular}%
}
\end{table*}

Sandeep et al. \cite{b45}evaluate four popular convolutional neural network architectures; ResNet152, VGG19, DenseNet169, MobileNetv3. The models were trained and validated using images of a publicly available brain MRI dataset from Kaggle which had 7,023 images classified into four classes; glioma, meningioma, pituitary tumors and normal brain images. To help with the imbalance class, a public dad set, and improve accuracy of the model, the researchers send data augmentation techniques such as rotation, zooming and flipping. The dataset was split into 80\% - 20\% of training to testing data set, and a five-fold validation was done to allow for the models to be tested for robustness. Out of all four models tested the MobileNetv3 model had the highest classification accuracy at 99.75\%, which was higher than the ResNet152, VGG19, or DenseNet169 architectures.

Zahoor \& Saddam \cite{b46}proposed a new deep learning framework called Res-BRNet which is used to improve brain tumor classification from MRI images. The Res-BRNet model addresses some of the unique problems associated with brain tumors, which has more complex shapes and varying texture, size, and location by using regional and boundary-based structures in spatial and residual blocks. The Res-BRNet model is structured so that the spatial blocks develop features related to tumor heterogeneity, homogeneity, and tumor boundaries (i.e., patterns and edges related to tumors). The residual blocks then learn local and global variations in texture (i.e., low-level features and high-level features) without having to specify the desired features of interest. This model design allows the Res-BRNet model to learn unique discriminative features needed for classification. The Res-BRNet model showed better performance than well-established CNN models with an accuracy of 98.22\%, sensitivity of 0.9811, precision of 0.9822, and F1 score of 0.9841.

Three various feature extraction methods were applied by Kumar et al.: Local Binary Patterns (LBP), Histogram of Oriented Gradients (HOG), and Image Loading\cite{b47}. These techniques were implemented in two publicly available MRI datasets of brain tumors examined in Kaggle. Six machine learning algorithms were used after feature extraction: Random Forest, Support Vector Machine (SVM), Logistic Regression, K-Nearest Neighbors (KNN), Naive Bayes, and Decision Tree. The results of the study showed that the Random Forest classifier had the highest classification accuracy of 99\% for brain tumors when paired with the Image Loading feature extraction method. SVM and Logistic Regression also performed well, though marginally worse than Random Forest, but this combination outperformed the others. Conversely, the KNN, Naive Bayes, and Decision Tree algorithms exhibited comparatively lower accuracy rates, underscoring the significance of selecting appropriate algorithm-feature extraction pairings for optimal classification results. 

The authors launch the Rotation Invariant Vision Transformer (RViT), a new type of deep learning model designed for classifying brain tumors found on MRI scans\cite{b48}. The main contribution of RViT is in the way it creates the rotated variants of the input images, after which each rotated image has been produced at various angles. Afterwards, the process of generating the patches and projecting them linearly into an embedding space continues. All rotated images create embeddings, which means that there are embeddings for all the rotated versions of the original MRI scans. RViT calculates a mean value of all embeddings and forms a rotation invariant representation of the original image. The strength of this process relates to the architecture learning features invariant to possible rotations, which then assists the model if the tumor is positioned differently in each scan. Overall, the authors achieved good accuracy measures including a sensitivity of 1.0, specificity of 0.975, F1-score of 0.984, Matthew's Correlation Coefficient (MCC) of 0.972, and overall accuracy of 0.986.

Our Xception model achieves the best training accuracy in the same dataset, 99.11\%, with a 0.0270 training loss. Still, our uniqueness is that our model not only gives us the output but also shows us how much data it detects as incorrect and the percentage of the incorrect data according to validation loss, which does not apply in other papers.

\section{Limitations and Future Work}
We should be more dependable in pre-trained models, although this model gives us better accuracy than any other custom models, as our custom CNN model didn’t perform well here. So, we need more fine-tuning so that we can build a more efficient Custom CNN model, which will perform with better accuracy and real-life applicability. We should make a smaller model for brain tumor detection. So, it can be easily used in real life, not only for internship doctors but also for other professional doctors to ensure the actual problems of their patients.

\section{Conclusion}
It investigates this field to shed further light on the application of deep learning mechanics in the automatic detection and classification of brain tumors in MRI images. Several pre-trained models were compared, and many custom architectures were proposed, with the Xception exception model being the most successful for its improved accuracy, less validation loss, and good cases of generalization. Highlighted are the essential roles played by AI-driven applications towards diagnostic effectiveness, clinical streamlining, and efficacy care in brain-cancer diagnosis.

Nevertheless, the present analysis is accepted to contain several limitations in that some models were computationally too heavy for demonstration and the variety of MRI images used was relatively small. Future work should target the frontend of development for light-weight scaling models using a high degree of accuracy while being deployable within a limited computing resource environment. Running experiments on different imaging formats can provide soundness for accuracy detection and segmentation via the syntheses of multi-modality data.

Broadly, the present research enforces the significancy of AI in medical imaging and its significancy regarding the solution of global issues in brain tumor detection and its treatment. Interfacing technological innovations with clinical practices, this study becomes a basal for a leap of advancement in automating neuroimaging diagnostics.

\vspace{12pt}
\color{red}

\end{document}